# The syntax-lexicon tradeoff in writing

Neguine Rezaii

Frontotemporal Disorders Unit, Department of Neurology, Massachusetts General Hospital, Harvard Medical School, Boston

Neguine Rezaii
nrezaii@mgh.harvard.edu
Frontotemporal Disorders Unit, Department of Neurology, Massachusetts General Hospital
149 13th Street, Suite 10.011, Charlestown, MA 02129, USA

**Abstract**

As speakers turn their thoughts into sentences, they maintain a balance between the complexity of words and syntax. However, it is unclear whether this syntax-lexicon tradeoff is unique to the spoken language production that is under the pressure of rapid online processing. Alternatively, it is possible that the tradeoff is a basic property of language irrespective of the modality of production. This work evaluates the relationship between the complexity of words and syntactic rules in the written language of neurotypical individuals on three different topics. We found that similar to speaking, constructing sentences in writing involves a tradeoff between the complexity of the lexical and syntactic items. We also show that the reduced online processing demands during writing allows for retrieving more complex words at the cost of incorporating simpler syntax. This work further highlights the role of accessibility of the elements of a sentence as the driving force in the emergence of the syntax-lexicon tradeoff.

# Introduction

Spoken language production occurs under the pressure of rapid online processing (1). This process involves retrieving both lexical and syntactic items with more complex items requiring more cognitive effort (2–6). Fulfilling the goal of communicating a clear message together with various time and cognitive constraints have rendered language its specific structure (7). Along the lines of what properties of language promote efficient communication, a recent work has shown that speakers maintain a balance between the complexity of lexical and syntactic items in their sentences (8). That is, a spoken sentence that contains complex words is more likely to have simpler syntax. Similarly, a sentence with complex syntax likely contains simpler words. To approximate the lexical and syntactic complexity, the work used a metric that was general enough to be applied to both items, that is the frequency of words and syntactic rules (9). The primary concern of the current work is to evaluate whether the syntax-lexicon tradeoff also exists in writing, a modality of language production that is less pressured by the online processing demands.

Unlike speakers, writers generally have more time and are cognitively freer to construct sentences allowing them to better plan and monitor their produced language (1, 10). It has been shown that writing is more structurally complex as measured by longer sentences, number of T-units, and a higher rate of subordination (11–15). With respect to the content, writing also allows for the inclusion of a higher rate of new ideas within a single idea unit than speaking (16). Written sentences have more content words and more complex noun phrases (17). As both lexical and syntactic items are more complex in written than spoken language, the syntax-lexicon tradeoff may not emerge in the written sentences, making it a feature unique to spoken language production.

This work probes the written language samples of neurotypical individuals for a possible syntax-lexicon tradeoff. First, we ask two independent cohorts of participants to describe a picture, one in writing and one in speaking, to evaluate the complexity of words and syntax and their relationship across the two modalities. We then examine the syntax-lexicon tradeoff in another language production elicitation task that involves the description of a job. Recently, it was shown that the syntax-lexicon tradeoff exists in the spoken samples of the picnic description task as well as a large corpus of language samples, but not in the description of one's job. Here, we hypothesize that the absence of this effect in the description of one's job stems from the fact that a corpus-based measurement of word frequency does not capture the idiosyncratic differences in what set of words are truly infrequent at an individual level. For example, in the description of their job, a professor of linguistics my say, "As a linguist, I am interested in long distance dependency relations", a sentence that contains both low-frequency words and low frequency syntax which makes it a counter example for the syntax-lexicon tradeoff. However, for the professor who frequently uses the term "dependency", this word does not impose a lexical access difficulty, sparing enough cognitive resources to retrieve low frequency syntactic rules. To test this hypothesis, we asked a separate cohort of neurotypical individuals to first write about a typical day in their job and then a typical day in the job of an archaeologist. We selected archaeology as it is a fairly uncommon job likely associated with words that are infrequently used by a typical person. In this context, we expect the syntax lexicon tradeoff as measured by a corpus-based measurement of frequency to be detected in the description of the typical day of an archaeologist but not one's own job.

# Methods

Written samples for the picture description task. For this task, we recruited 83 participants from Amazon's Mechanical Turk (MTurk). MTurk participants filled out the short and validated version of the Everyday Cognition test with twelve items (ECog-12), an informant-rated questionnaire designed to detect cognitive and functional decline (18). Only language samples from participants who were native English speakers, with no self-reported history of brain injury or speech/language disorder, either developmental or acquired, were included in the analyses. The participants were asked to look at a drawing of a family at a picnic from the Western Aphasia Battery–Revised (19) and describe it using as many full sentences as they could.

Spoken samples for the picture description task. Thirty-five healthy speaking participants were enrolled through the Speech and Feeding Disorders Laboratory at the Massachusetts General Hospital (MGH) Institute of Health Professions. These participants passed a cognitive screen, were native English speakers, and had no history of neurologic injury or developmental speech/language disorders.

Written samples for the job description task. In a two-step paradigm, we asked a separate cohort of 54 MTurk participants to first describe a typical day of their most recent job and then the typical day in the work of an archaeologist using as many full sentences as possible. The procedure for recruiting the participants was the same as the one used for the written picture description task.

Text analysis of language samples. We used Quantitext, a text analysis toolbox we developed in Frontotemporal Disorders Unit of Massachusetts General Hospital, to automatically produce a set of quantitative language metrics. The goal of developing this package is to increase the precision and objectivity of language assessments while reducing human labor (as outlined in (20)). The toolbox uses a number of natural language processing toolkits and software such as Stanford Parser (21), spaCy (22), and NLTK as well as text analysis libraries in R. Quantitext receives transcribed language samples as input and generates as outputs a number of metrics such as sentence length, log word frequency, log syntax frequency (8, 9), content units (as in (23)), efficiency of lexical and syntactic items (as in (24)), part of speech tags, and the distinction of heavy and light verbs.

Measuring the frequencies of syntactic rules and words. To measure syntactic rule and word frequencies, we used the Switchboard corpus (25), which consists of spontaneous telephone conversations averaging 6 minutes in length spoken by over 500 speakers of both sexes from a variety of dialects of American English. We use it to estimate word and syntax frequencies in spoken English, independently of our patient and control sample. The corpus contains 2,345,269 words. Content word frequencies reported in this study (which we use as our main measure of word frequency) are based on frequencies of main verbs (excluding auxiliary verbs, and *be*, *do*, and *have*), nouns, adjectives, and adverbs.

Statistics. For the statistical analyses of this study, we used the R software version 4.1.2. To evaluate the syntax-lexicon tradeoff, we used mixed-effects models with subject-specific random intercept via the lme4 package in R (26).

## Results

1. Comparing written and spoken in the picture description task

The average sentence length in the written picture description 8.739394 (SD = 3.99). The sentences had an average content word frequency (log) of 5.18 (SD = 1.11), average all word frequency (log) of 7.76 (0.81), and average syntax frequency (log) of 7.40 (SD = 1.56).

To compare written and spoken language samples, we fit a mixed effects model with subject-specific random intercept to predict the metrics of sentence length, content word frequency, all word frequency and syntax frequency at the sentence level with the modality of language production as a predictor. A separate model was run for each of these metrics. For models on content word frequency, all word frequency and syntax frequency, we also included sentence length as a predictor. Compared to spoken sentence in the picture description task, written sentences contained lower content word frequency ($\beta$=-0.230, SE= 0.109, t = -2.103, p = 0.039), lower all word frequency ($\beta$=-0.367, SE= 0.071, t = -5.194, p < 0.001), and higher syntax frequency ($\beta$= 0.316, SE = 0.137, t = 2.300, p = 0.024). We did not detect a significant difference in the sentence length between the written and spoken sentences ($\beta$=-0.703, SE= 0.465, t = -1.507, p = 0.136).

2. The syntax-lexicon tradeoff in the written sentences of the picture description task

To evaluate the syntax-lexicon tradeoff at the sentence level, we fit a mixed effect model that predicts content word frequency, with syntax frequency and sentence length as predictors and random intercepts for subject, we found a main effect of syntax frequency ($\beta$ = -0.169, SE = 0.041, t = -4.168, p = 0.001) and sentence length ($\beta$=-0.028, SE= 0.014, t = -2.058, p = 0.040)[1].

Combining all written and spoken sentences in the picture description task as shown in Figure 1, we ran a regression predicting word frequency from syntax frequency, whether the modality was written versus spoken, and the interaction of modality and syntax frequency. We included a random intercept for subject but no random slopes to aid convergence. We found that the modality of language production did not significantly moderate the effect of syntax frequency on word frequency, as evidenced by a nonsignificant interaction term (for the interaction term, $\beta$ = −0.005, P = 0.924).

---

[1] Similarly, fitting a mixed effect model that predicts all word frequency, with syntax frequency and sentence length as predictors and random intercepts for subject (with a random slope for syntax frequency but no correlations and no random slopes for length, in order to aid convergence), we found a main effect of syntax frequency ($\beta$ = -0.068, SE = 0.026, t = -2.579, p = 0.011), but not sentence length ($\beta$ = 0.014, SE = 0.010, t = 1.422, p = 0.156).

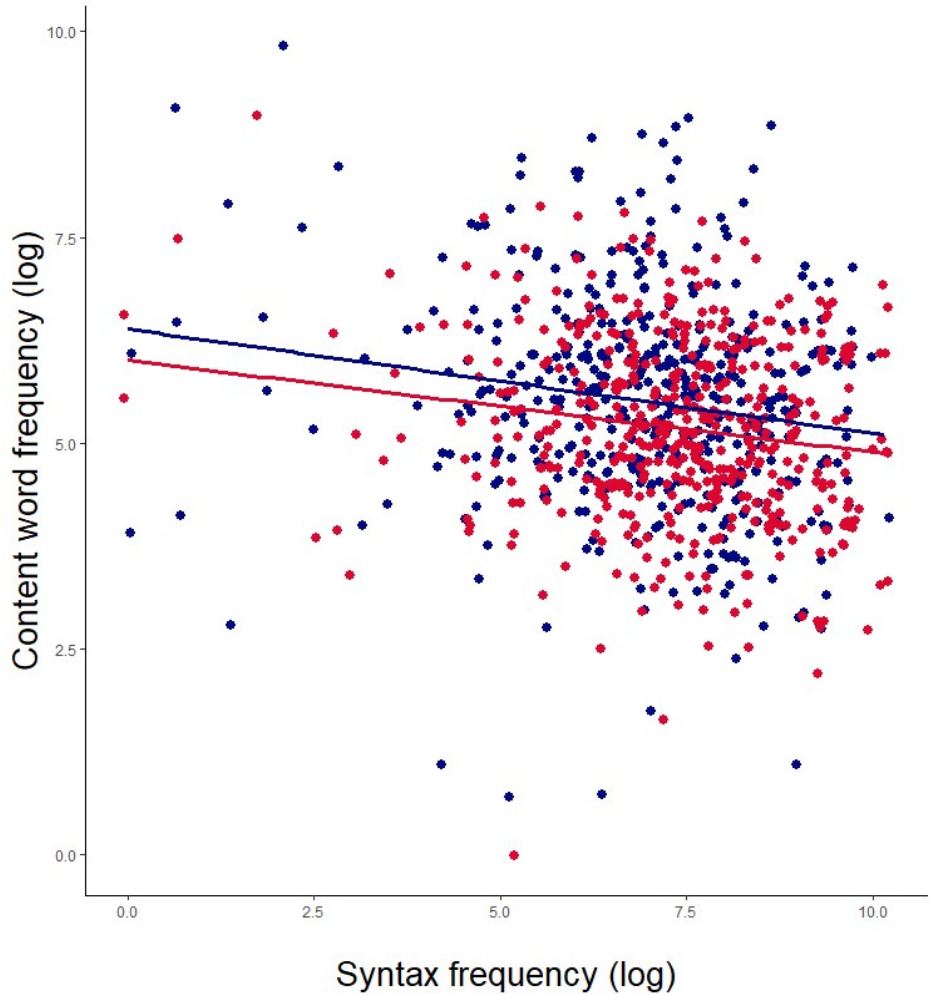

**Figure 1.** The syntax lexicon tradeoff in the picture description task in spoken (dark blue line) and written samples (red line)

3. Comparing written sentences of the description of one's own job versus an uncommon job (archaeology)

    To compare the written sentences of one's own job versus the job of an archaeologist, we fit a mixed effects model with subject-specific random intercept to predict the metrics of sentence length, content word frequency, all word frequency and syntax frequency at the sentence level with the modality of language production as a predictor. A separate model was run for each of these metrics. For models on content word frequency, all word frequency and syntax frequency, we also included sentence length as a predictor. We did not find a difference in content word frequency ($\beta$ = 0.247, SE = 0.147, t = 1.685, p = 0.093), all word frequency ($\beta$ = 0.065, SE = 0.179, t = 0.544, p = 0.587), and syntax frequency ($\beta$ = -0.222, SE = 0.158, t = -1.402, p = 0.162) between the two topics. The written sentences on one's own job were shorter than on the job of an archaeologist ($\beta$ = -2.018, SE = 0.811, t = -2.488, p = 0.014).

4. The syntax-lexicon tradeoff in the written sentences of the job description task

    To evaluate the syntax-lexicon tradeoff in the written sentences of one's work, we fit a mixed effect model that predicts content word frequency, with syntax frequency and sentence length as predictors and random intercepts for subject (with a random slope for syntax frequency but no correlations and no random slopes for length, in order to aid convergence), we did not find a main effect

of syntax frequency (β = -0.023, SE = 0.572, t = -0.287, p = 0.776), nor of sentence length (β = -0.012, SE= 0.015, t = -0.745, p = 0.540). [2]

In the written sentences on the job of an Archaeologist, we fit a mixed effect model that predicts content word frequency, with syntax frequency and sentence length as predictors and random intercepts for subject (with a random slope for syntax frequency but no correlations and no random slopes for length, in order to aid convergence), we found a main effect of syntax frequency (β = -0.269, SE = 0.093, t = -2.877, p = 0.005), but not sentence length (β = -0.026, SE = 0.017, t = -1.596, p = 0.114)[3] Figure 2.

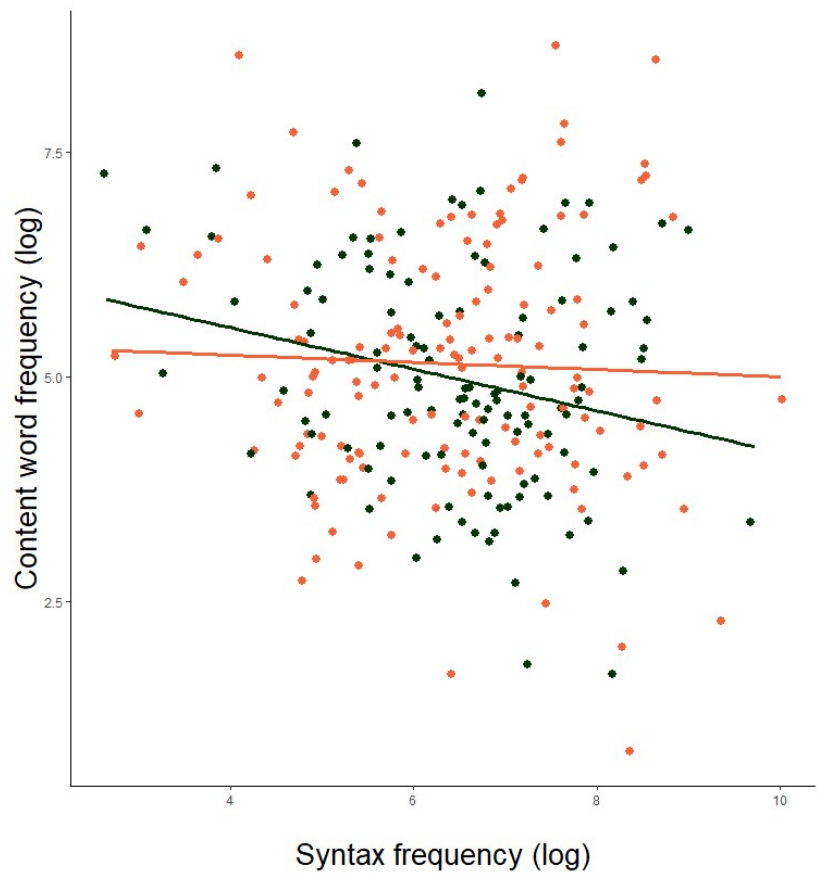

**Figure 2.** The syntax-lexicon tradeoff of the written sentences in the description of one's own job (orange line) versus an archaeologist's job (dark green line)

---

[2] Similarly, fitting a mixed effect model that predicts all word frequency, with syntax frequency and sentence length as predictors and random intercepts for subject (with a random slope for syntax frequency but no correlations and no random slopes for length, in order to aid convergence), we did not find a main effect of syntax frequency (β = -0.054, SE = 0.442, t = -0.785, p = 0.436), nor of sentence length (β = -0.015, SE = 0.011, t = -1.341, p = 0.182).

[3] Similarly, fitting a mixed effect model that predicts all word frequency, with syntax frequency and sentence length as predictors and random intercepts for subject (with a random slope for syntax frequency but no correlations and no random slopes for length, in order to aid convergence), we found a main effect of syntax frequency (β = -0.370, SE = 0.119, t =-3.097, p = 0.012) but not sentence length (β = -0.037, SE = 0.020, t = -1.800, p = 0.076).

# Discussion

In this work, we showed that that the syntax-lexicon tradeoff that has been recently reported in spoken language also exists in written samples, suggesting it to be a general property of language independent of the modality of production.

In the cognitive neuroscience of language production, there has been a disagreement about extent to which written language is supported by the same cognitive and neuronal systems as spoken language (27). Many studies have emphasized the distinction between speaking and writing (10) by showing that written language contains more complex structures, more content words, more complete idea units and higher concentration of new information than speech (12, 14–16, 28–31). In a dual system theory for writing and speaking, the co-occurrence of written and spoken deficits is considered to be coincidental and a result of simultaneous damage to the brain regions that support each function (32–34). Other studies, on the other hand, argue that the skills needed for literacy such as reading and writing likely have no neuronally evolved systems in the brain (35). As a result, these skills are built upon the naturally evolved neural mechanisms for speaking and comprehension. This argument, motivated by connectionist models of language processing, is supported by evidence showing that deficits in writing and reading arise because of damage to central semantic and phonological processes that support language production and comprehension (36–38). This hypothesis further supports the ease with which the majority of school children learn the letter-sound correspondence in a few months, despite the fact that the brain is not phylogenetically evolved to develop written language skills (39).

The comparison between spoken and written samples in the description of a picture of this work showed that written sentences contain lower frequency words but higher frequency syntax with no difference in the sentence length. Despite the differences in the average complexity of lexical and syntactic items across speaking and writing, there continues to exist a balance between the complexity of syntax and lexicon. This finding further suggests that the more timing and planning that are available for writing are mainly allocated to the retrieval of more complex words at the cost of using simplifying the syntax. The existence of the syntax-lexicon tradeoff in both writing and speaking suggest that at least some functions lie at the core of language production regardless of the specific modality (38).

This work further showed that the syntax-lexicon tradeoff would emerge when there is an access difficulty in retrieving either lexical or syntactic items to overwhelm the channel capacity of language production. Although in most cases, a corpus-based measurement of frequency of the lexical and syntactic reflects an access difficulty for an individual (40, 41), it does not capture the inter-individual variability in familiarity or exposure with certain words (or syntax). The lexicon used to describe one's job is an example of such variability as each individual is routinely exposed with certain words that are not necessarily frequent in a large corpus of average word use. In this case, a corpus-based measurement of word frequency would be a false metric for access difficulty. Future work is needed to explore the ways in which, the measures of lexical and syntactic accessibility could be individualized.

## Acknowledgements

We thank Dr. Jordan Green for sharing speech samples from control participants.